\documentclass[10pt, conference]{ieeeconf}

\usepackage{amsmath,amssymb,amsfonts}
\usepackage{graphicx, stfloats}
\usepackage{pifont}

\usepackage{xcolor}
\usepackage[figuresright]{rotating}

\graphicspath{{/}{fig/}}

\usepackage{array}
\usepackage{textcomp}
\usepackage{xcolor}
\usepackage{multirow}
\usepackage{booktabs}

\usepackage{mathtools}
\usepackage{breqn}
\usepackage{float}

\usepackage{pgfplots}
\pgfplotsset{compat=1.7}
\usepgfplotslibrary{groupplots}

\usepackage{caption}
\usepackage{subcaption}

\usepackage{multirow,tabularx}
\usepackage{hyperref}
\usepackage{flushend}
\usepackage{algorithmic}
\usepackage[vlined, ruled, shortend]{algorithm2e}

\usepackage{multicol}
\usepackage{times}
\usepackage{cleveref}
\usepackage{soul}
\DeclareMathOperator*{\argmin}{arg\,min}

\usepackage{cuted}

\newlength\figureheight
\newlength\figurewidth
\SetAlCapNameFnt{\footnotesize}
\SetAlCapFnt{\footnotesize}

\newcommand{\ie}{\textit{i.e.,}}%
\newcommand{\eg}{\textit{e.g.,}}%
\newcommand{\etal}{\textit{et al.}}%

\crefname{figure}{Fig.}{Figs.}
\Crefname{figure}{Fig.}{Figs.}
\Crefname{section}{Section}{Section}
\crefname{section}{Section}{Section}

\usepackage{etoolbox}

\title{
    \Huge
    Aion: Towards Hierarchical 4D Scene Graphs with Temporal Flow Dynamics
}

\author{
    Iacopo Catalano, Eduardo Montijano, Javier Civera, Julio A. Placed\authorrefmark{2}, Jorge Peña-Queralta\authorrefmark{2}
    \thanks{\authorrefmark{2} Equal Project Management.}
    \thanks{This work was partially supported by the Kaute Foundation through the Tutkijat Maailmalle program, by DGA\_FSE T73\_23R and by project UNDERAIBOT (CPP2022-009792) funded by MICIU/AEI/10.13039/501100011033 and European Union (NextGenerationEU/PRTR).}
    \thanks{Iacopo Catalano is with the University of Turku, 20014, Turku, Finland. (e-mail: imcata@utu.fi).}
    \thanks{Jorge Pe\~na-Queralta is with the Centre for Artificial Intelligence, Zürich University of Applied Sciences, Winterthur, Switzerland. (e-mail: penq@zhaw.ch).}
    \thanks{Julio A.~Placed is with the Instituto Tecnol\'ogico de Arag\'on (ITA) and the University of Zaragoza, Mar\'ia de Luna 3-7, Zaragoza, Spain (e-mail: jplaced@ita.es).}
    \thanks{Javier Civera and Eduardo Montijano are with the University of Zaragoza, 50018, Zaragoza, Spain. (e-mail: jcivera@unizar.es, emonti@unizar.es).}
}

\begin{document}
\IEEEoverridecommandlockouts

\maketitle

\begin{strip}
    \begin{minipage}{0.33\textwidth}
        \vspace{-10pt}
        \centering
        \includegraphics[width=\textwidth]{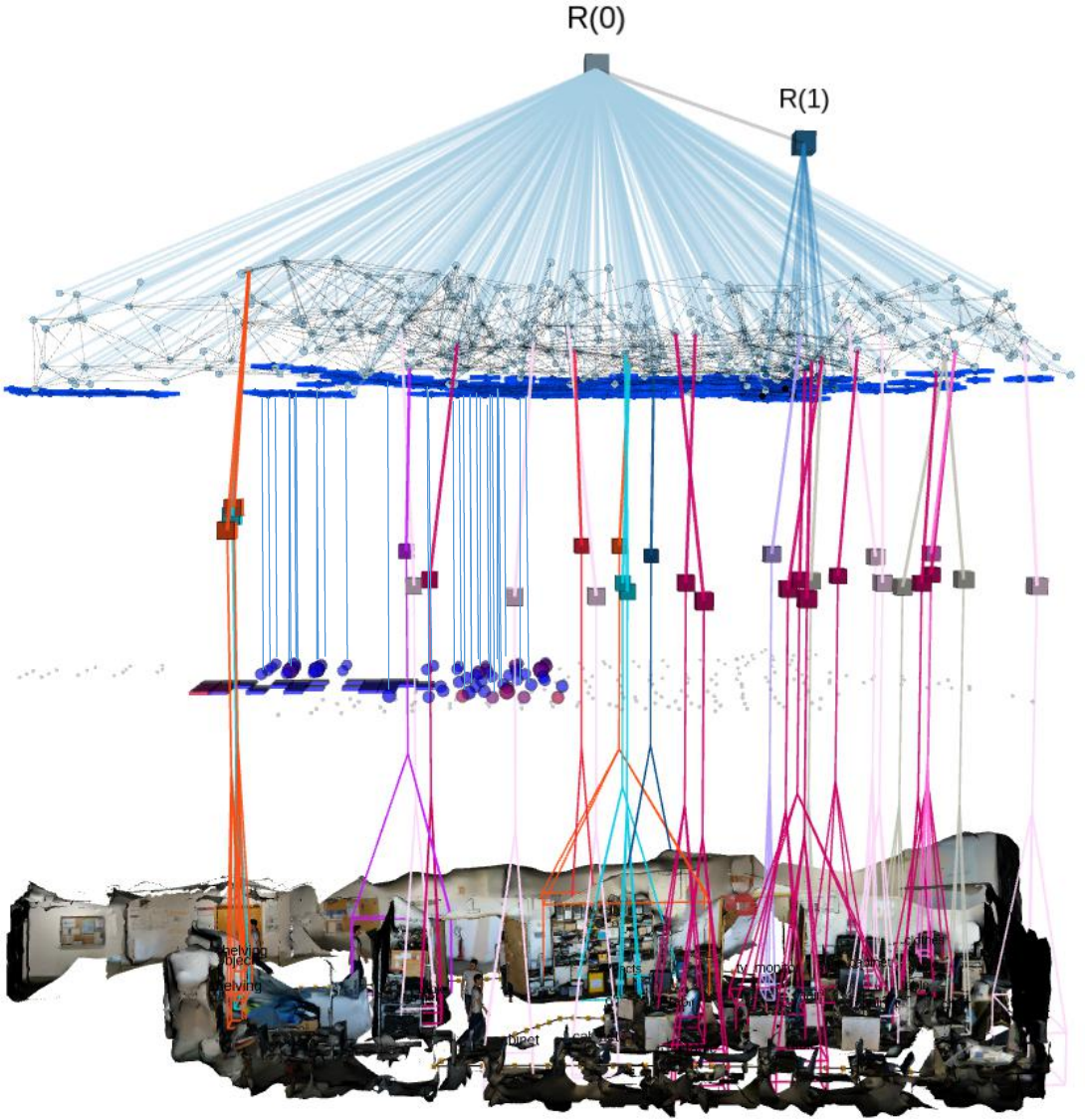} \\
        {\small (a) Temporal 3D scene graph.}
        \label{fig:aion-full}
    \end{minipage}%
    \hfill
    \begin{minipage}{0.65\textwidth}
        \centering
        \includegraphics[width=\textwidth]{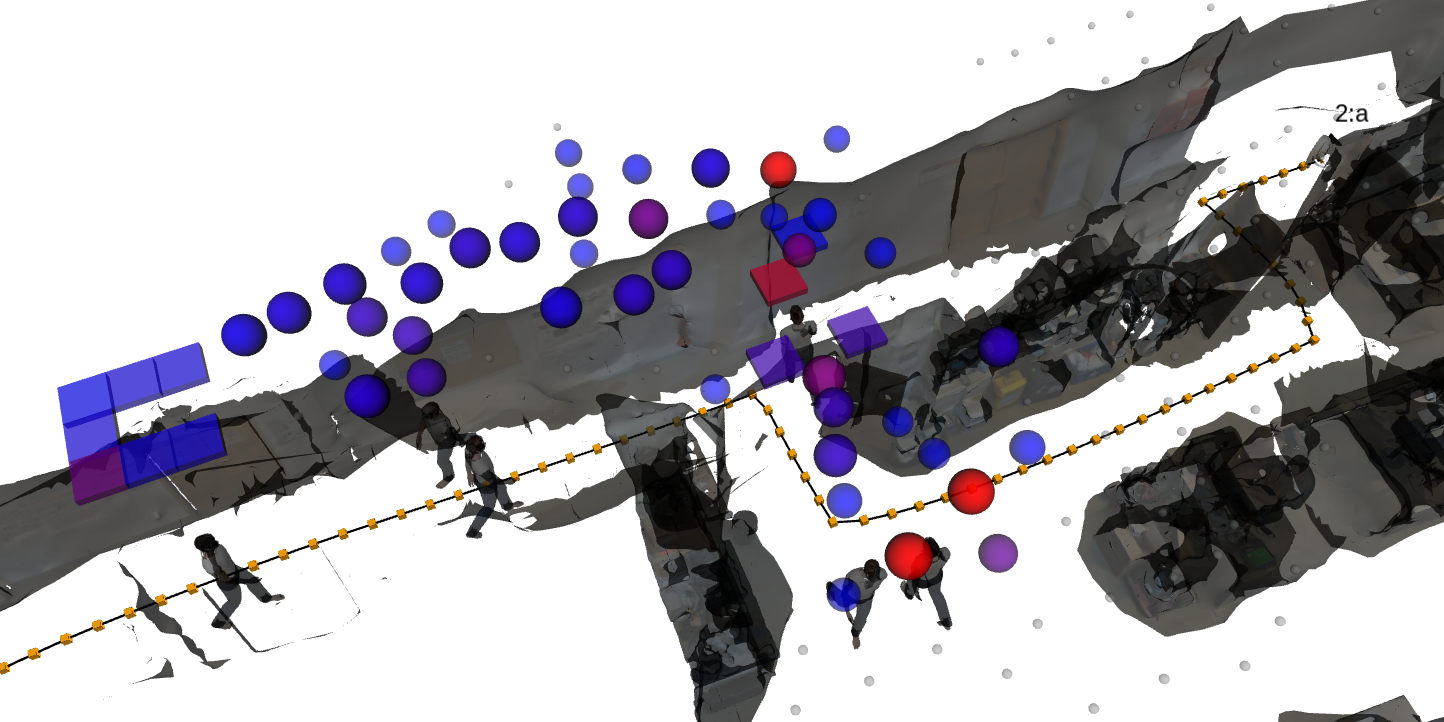} \\
        {\small (b) Temporal flow dynamics layer.}
        \label{fig:temporal-dynamics}
    \end{minipage}
    \captionof{figure}{We propose {Aion}, a framework for modeling temporal flow dynamics within hierarchical spatial representations by integrating frequency-domain temporal flow modeling. The system transforms static 3D Scene Graphs (a) into 4D spatio-temporal representations, where navigational nodes are augmented with temporal flow predictions (b). Flow descriptors are visualized as spheres, where size encodes flow magnitude and color encodes directional entropy (blue: low entropy, red: high entropy). Flat square represent hash cells containing unbound dynamics.}
    \label{fig:concept}
\end{strip}

\begin{abstract}
Autonomous navigation in dynamic environments requires spatial representations that capture both semantic structure and temporal evolution. 
3D Scene Graphs (3DSGs) provide hierarchical multi-resolution abstractions that encode geometry and semantics, but existing extensions toward dynamics largely focus on individual objects or agents. 
In parallel, Maps of Dynamics (MoDs) model typical motion patterns and temporal regularities, yet are usually tied to grid-based discretizations that lack semantic awareness and do not scale well to large environments. 
In this paper we introduce \textbf{Aion}, a framework that embeds \emph{temporal flow dynamics} directly within a hierarchical 3DSG, effectively incorporating the temporal dimension. 
Aion employs a graph-based sparse MoD representation to capture motion flows over arbitrary time intervals and attaches them to navigational nodes in the scene graph, yielding more interpretable and scalable predictions that improve planning and interaction in complex dynamic environments. We provide the code at \href{https://github.com/IacopomC/aion}{github.com/IacopomC/aion}

\end{abstract}
\IEEEpeerreviewmaketitle


\section{Introduction}\label{sec:introduction}

Autonomous robots operating in human-populated environments must navigate complex dynamic scenes where understanding spatial structure alone is insufficient for safe and efficient operation~\cite{grzeskowiak2021crowd}. Anticipating human movements and environmental changes is critical in virtually any operating environment, not only to avoid collisions but also to plan proactively accounting for past history or predicted motions~\cite{paez2022pedestrian}.

Recently, 3D Scene Graphs (3DSG) have emerged as powerful abstractions for robotic spatial understanding, encoding hierarchical semantic structures that capture both geometric and semantic relationships in environments~\cite{catalano20253d}. These layered graphs encode geometric, semantic, and topological information at multiple levels of abstraction (\eg~low-level mesh geometry, high-level room semantics) providing a natural discretization of space into semantically meaningful locations. However, existing 3DSGs are fundamentally static, modeling spatial structure at a time instant and lacking the ability to capture and predict temporal flow dynamics, critical for autonomous systems operating in populated, changing environments. A number of works have introduced dynamics by modeling the temporal position of movement of either objects or agents (\eg~human actors)~\cite{rosinol2020dsg, greve2023curb}. These approaches, while valuable in many environments, are not applicable to global dynamics.

In the domain of temporal flow modeling, recent research on Maps of Dynamics (MoDs)~\cite{kucner2023survey} introduced models encoding spatiotemporal motion patterns, enabling robots to predict and reason about future environment states.
Prior MoD approaches have predominantly focused on embedding dynamics in uniform grid-based occupancy maps, applying spectral and probabilistic methods to learn temporal patterns~\cite{krajnik2014spectral, kucner2017enabling, molina2018modelling}. While effective, these grid-based models lack semantic understanding, enforce uniform spatial discretization regardless of place importance, and cannot leverage the hierarchical spatial abstractions that humans naturally use to navigate complex environments. 
Finally, these approaches do not scale well with the size of the environment, as hierarchical representations do.

This paper introduces \textbf{Aion}, a system that integrates temporal flow dynamics directly into hierarchical 3D scene graphs, bridging the gap between rich spatial-semantic, scalable representations and predictive temporal flow modeling. Unlike prior MoD approaches that focus on grid cells, Aion learns and embeds temporal patterns at the navigational level by leveraging the natural hierarchical discretization of 3DSGs. This enables robots to generate more interpretable and actionable predictions of how specific parts of the environment evolve over time, improving navigation and interaction in dynamic human environments. Aion represents the first 3D extension of MoDs within a semantically informed scene graph framework. Through the integration of temporal flow dynamics, we bridge the gap toward a Hierarchical 4D Scene Graph.

To deliver the above core contributions of Aion, this paper introduces:
\begin{itemize}
    \item \textbf{Graph-based Maps of Dynamics:} We extend the concept of MoDs from uniform grid-based to semantically-informed hierarchical spatial representations, enabling temporal reasoning at meaningful navigation locations rather than arbitrary spatial discretizations.

    \item \textbf{Dynamic Topology Temporal Modeling}: We solve the fundamental challenge of maintaining temporal model consistency over dynamic scene graph topologies through position-invariant indexing, enabling seamless temporal learning as spatial understanding evolves during exploration (\eg~in the event of loop closures).

    \item \textbf{3DSG Integration and Flow Prediction:} Our approach enables temporal flow prediction directly for navigational nodes (through integration in Hydra~\cite{hughes2022hydra}) rather than arbitrary spatial grid cells, providing more interpretable and actionable temporal information for navigation planning.
\end{itemize}

\section{Related Work}\label{sec:related}

\subsection{3D Scene Graphs (3DSGs)}

Hierarchical 3DSGs are structured representations that integrate geometric and semantic information to support spatial understanding in robotic systems~\cite{catalano20253d}. By modeling environments as layered graphs 3DSGs enable reasoning and decision-making across multiple levels of abstraction (from low-level geometry to semantically meaningful entities such as objects, rooms, and buildings) enabling reasoning across multiple spatial and semantic scales~\cite{chang2023hydra, hughes2024foundations, bavle2023s}. This hierarchical structure allows robots to efficiently interpret, plan, and act in complex environments by leveraging compact semantic concepts instead of dense geometric data~\cite{hughes2024foundations, ejaz2025situationally}. Extensions to urban~\cite{greve2023curb, deng2024opengraph} and agricultural~\cite{mukuddem2024osiris} domains further demonstrate the adaptability of these models to domain-specific structures.

While prior work captures static or incrementally updated structure, it largely assumes a fixed spatial topology and lacks models of temporal evolution or dynamic state within the 3DSG structure itself. Action-aware graphs~\cite{ravichandran2022hierarchical, agia2022taskography, looper20233d} introduce affordances into the hierarchy, but do not model how environments evolve over time. In contrast, learning-based 3DSGs~\cite{chen2024clip, wang2023vl, lv2024sgformer} operate mainly on flat, object-centric graphs and focus on perception or language grounding, often without structural abstraction or dynamics.

\textbf{Dynamic Scene Graphs:} Specific works have extended 3DSGs to handle dynamic environments. Rosinol \etal~\cite{rosinol2020dsg, rosinol2021kimera} introduce 3D Dynamic Scene Graphs to jointly represent geometry, objects and agents, capturing dynamic entities alongside static structure. Similarly, CURB-SG \cite{greve2023curb, steinke2025collaborative} incorporates dynamic vehicles in its structure for urban mapping.

These methods demonstrate the value of dynamic scene graphs, however, they primarily focus on modeling agent trajectories, object motion, or scene evolution at the instance level. Such approaches do not scale well when capturing object flows (\eg~how crowds navigate complex indoor environments such as airports or college campuses). 
In contrast, our work directly integrates \emph{temporal flow dynamics} into the hierarchical structure of 3DSGs, transforming them for the first time into 4DSGs. This enables predictive reasoning about how people collectively move through environments over time, going beyond the dynamics of individual objects or agents.

\subsection{Maps of Dynamics (MoDs) and Temporal Modeling}

MoDs aim to capture  typical patterns of motion in space, enabling robots to anticipate how agents or objects move through an environment~\cite{kucner2023survey}. Existing approaches fundamentally differ in what types of input data they require, and how they model spatial understanding. Early models~\cite{kucner2013conditional, wang2014modeling} use probabilistic frameworks to represent local transitions and directional flows between spatial regions, encoding movement patterns through structured dependencies. Subsequent extensions generalize these concepts by associating multimodal velocity distributions with discrete spatial locations, supporting richer motion representation based on long-term sensor data~\cite{kucner2017enabling, senanayake2017bayesian}. While effective for short-term flow estimation and reactive planning, these approaches often rely on grid-based abstractions and are limited in modeling long-term temporal variations.

To address long-term dynamics, other frameworks model environments as time-varying probabilistic processes that capture periodicities in agent behaviors or environmental states, such as daily or seasonal cycles~\cite{krajnik2017fremen, molina2018modelling, vintr2019time, krajnik2014spectral}. These models facilitate predictive reasoning beyond static maps by learning temporal regularities. Recent efforts~\cite{shi2023learning, shi2025learning} further extend this concept by learning how spatial motion patterns themselves evolve over time, by predicting them directly from egocentric views~\cite{catalano2026egomod} or by incorporating static environmental geometry to inform dynamic predictions~\cite{verdoja2024bayesian}.
Despite their advantages, existing MoDs typically operate independently of higher-level semantic and structural information. In contrast, our approach integrates temporal flow dynamics directly within a hierarchical 3DSG, enabling structured reasoning and predictive planning over dynamic environments.

\section{Method}\label{sec:methodology}

Aion addresses the core challenge of integrating temporal flow dynamics within hierarchical 3D scene graphs, yielding a 4D spatio-temporal representation that augments 3DSGs with time-dependent behavioral information, enabling consistent temporal state representation over dynamic graph structures.

\subsection{3D Scene Graph Structure}

\begin{figure}
    \centering
    \includegraphics[width=.85\columnwidth]{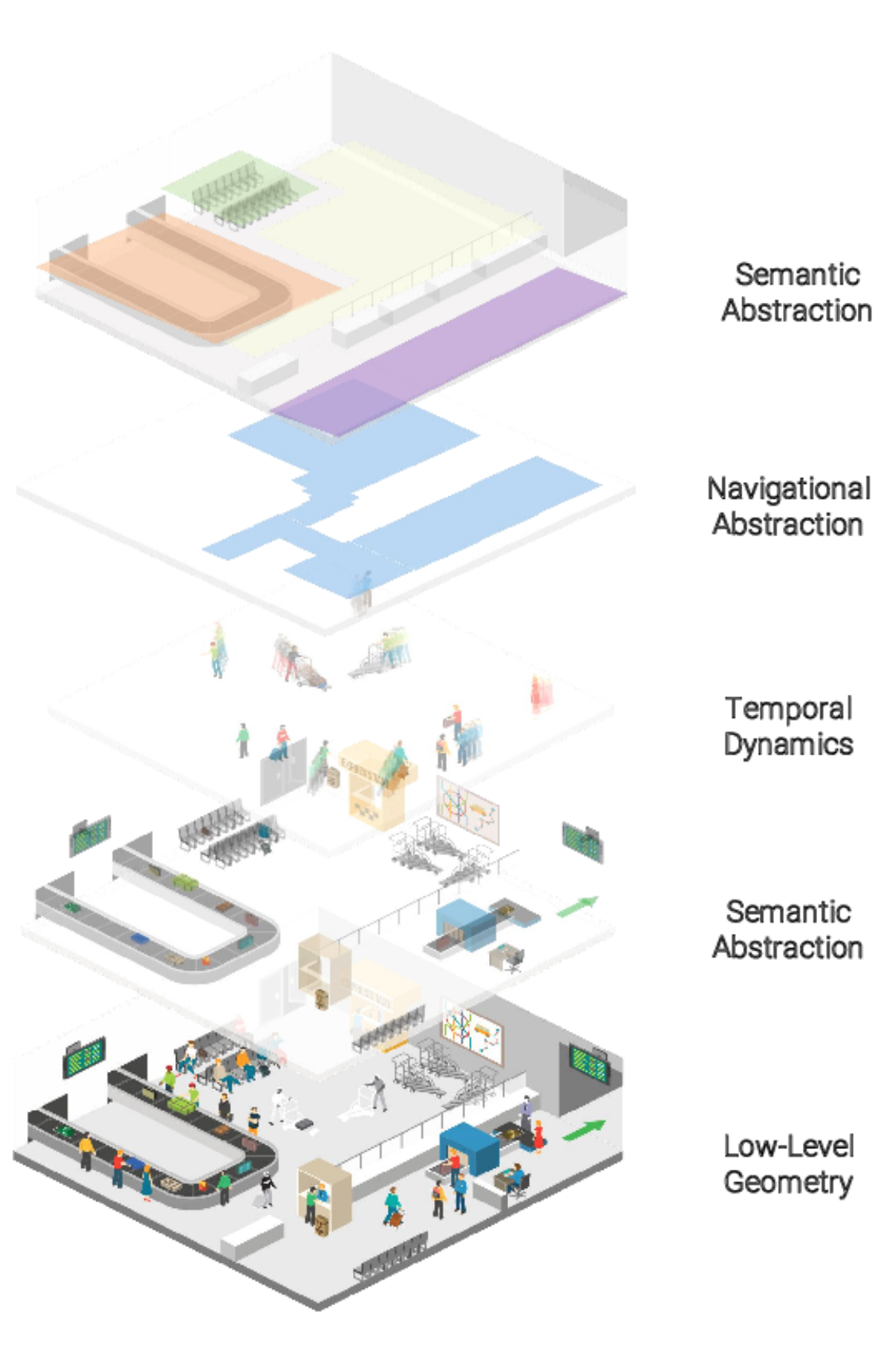}
    \caption{Illustration of the hierarchical decomposition of a scene into different geometric, semantic and navigational layers, where temporal flow dynamics emerge from semantic structure and play a central role to enriching navigation.}\label{fig:layers_3d}
\end{figure}

A 3DSG is a structure used to represent entities in a 3D environment along with their spatial and semantic relationships. At a given time step $t$, the scene is modeled as the graph {\small $\mathcal{G}^t \triangleq (\mathcal{V}^t, \mathcal{E}^t)$}, where nodes {\small $\mathcal{V}^t$} correspond to physical or conceptual elements, and edges {\small $\mathcal{E}^t$} encode their relational structure. If organized hierarchically, the node set {\small $\mathcal{V}^t$} is partitioned into disjoint subsets {\small $\mathcal{V}^t = \bigsqcup_{\ell=1}^{L} \mathcal{V}^t_\ell$} corresponding to different abstraction levels {\small $\ell \in \{1, \dots, L\}$}, allowing the graph to encode both fine-grained geometry and high-level semantic groupings. Nodes are grouped in this case across levels of abstraction as {\small $\boldsymbol{v}^t_{\ell,i} \in \mathcal{V}^t_\ell$} \cite{catalano20253d}.
Each layer in this hierarchy serves a distinct purpose (see~\cref{fig:layers_3d}):

\renewcommand{\theenumi}{\roman{enumi}}%
\begin{enumerate}
    \item {Low-Level geometry}: Captures the physical layout of the environment, \eg~through 
    meshes or point clouds.
    \item {Motion graph/Spatial anchoring}: Anchors observations in space and time, often encoding motion, agent trajectories, or dynamic entities.
    \item {Navigational abstractions and action affordances}: Represents traversable regions and their connectivity to support planning and action.
    \item {Semantic abstractions}: Groups entities into meaningful categories (\eg~objects) for higher-level understanding.
    \item {Global structure}: Integrates lower-level representations into a unified model of large-scale environments.
\end{enumerate}

This layered structure allows 3DSGs to support both low-level geometric reasoning and high-level semantic interpretation, enabling robust scene understanding across a range of tasks in robotics and embodied AI.

Our approach leverages the inherent hierarchy of the scene graph to model how activity patterns evolve over time, extending this framework to incorporate \emph{temporal flow dynamics} at the navigational level by adding an additional layer of abstraction to the conventional 3DSG scene decomposition (see~\cref{fig:layers_3d}). 

This temporal layer augments navigational nodes with directional flow statistics and predictive models, enabling the graph to capture not only where motion occurs but also how it changes over time. It provides the foundation for the methods introduced in the following sections, including our spatio-temporal modeling, sparse spatial hashing, and global temporal prediction mechanisms.

\subsection{Spatio-Temporal Modeling}

To model directional dynamics over time, Aion maintains sparse per-location orientation histograms, inspired by the grid-based motion modeling approach of~\cite{molina2018modelling} but extended to operate over navigational nodes in a sparse spatial hash. 
For notational simplicity, let us denote by {\small $\boldsymbol{v}^t_{n,i} \in \mathcal{V}^t_n$} the $i$-th node that belongs to the navigational abstraction layer {\small$\mathcal{V}^t_n$} at time $t$.
Each node maintains a time-varying activity vector representing directional motion, discretized into {\small$B$} angular bins over {\small$[0, 2\pi)$}. Observed orientations are mapped to bins and incrementally accumulated to form a historical activity profile per node. Formally, each navigational node $v^{t}_{n,i}$ is associated with a temporal activity vector:
\begin{equation}
    \boldsymbol{s}_i^{}(t) \in \mathbb{R}^{B\times\lambda} \,,
\end{equation}
\noindent where the $b$-th entry {\small$\boldsymbol{s}_{i,b}(t)\in\mathbb{R}^\lambda$} denotes the observed activity level in direction bin $b$ at time $t$, and {\small$\lambda$} is the dimension of the motion descriptor. The vector $\boldsymbol{s}_i^{}(t)$ represents motion directions in the global coordinate frame, ensuring that the activity history remains spatially meaningful even if nodes are repositioned.
The motion descriptors, such as flow magnitude, dominant direction, and directional entropy are computed from raw historical counts. These reflect empirical, observed motion patterns. Periodically, the histograms are normalized and passed to a global FreMEn model~\cite{krajnik2017fremen}, which captures temporal periodicity and enables prediction of future motion trends at each node. Predicted vectors retain the same angular structure but represent model-inferred directional likelihoods rather than raw counts.

\subsection{Sparse Spatial Hashing for Scalable Temporal Modeling}

\begin{figure}
    \centering
    \includegraphics[width=\columnwidth]{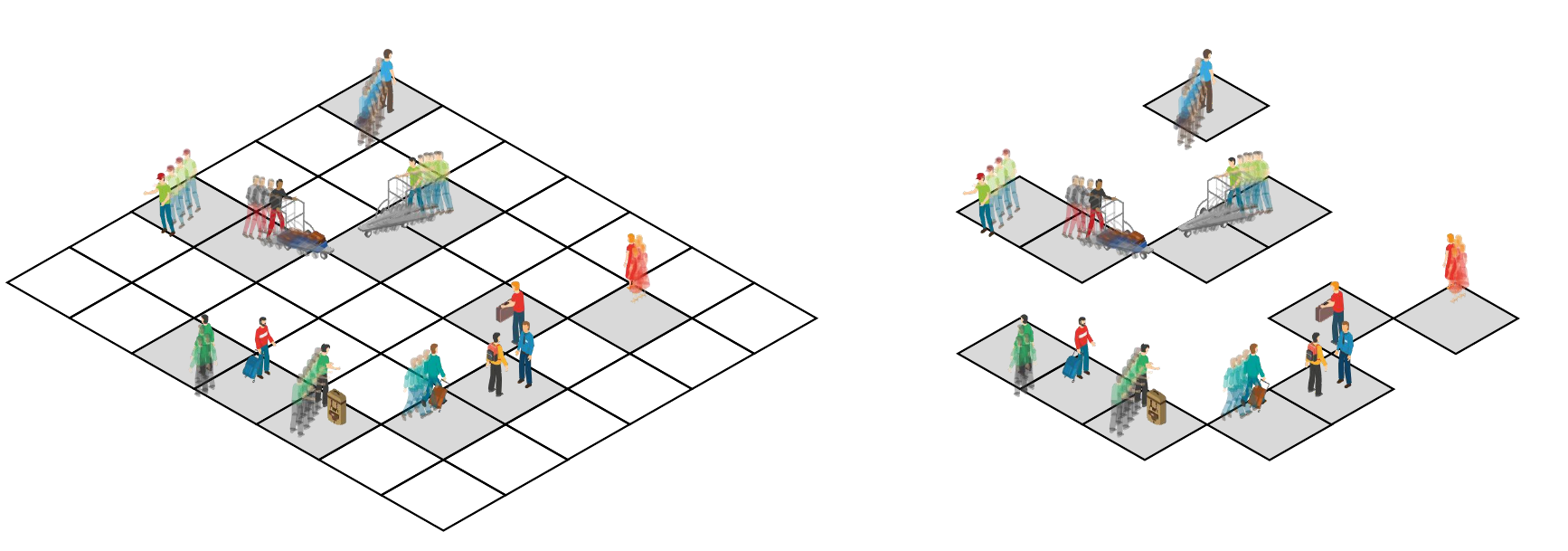}
    \caption{Comparison between grid-based models (left) and the proposed sparse spatial hashing of Aion (right) to encode temporal flow dynamics. While the grid cells represent the same spatial locations, only occupied cells are stored in memory in the case of Aion.}\label{fig:spatial_hashing}
\end{figure}

One of the key challenges is maintaining temporal model consistency as the set of navigational nodes evolves. Traditional grid-based temporal modeling approaches allocate memory for every spatial cell within predefined boundaries~\cite{molina2018modelling}, leading to computational overhead when applied to realistic environments where activity is spatially sparse. To overcome this limitation, Aion adopts a sparse spatial hashing mechanism (see~\cref{fig:spatial_hashing}) that enables infinite spatial coverage while maintaining $\mathcal{O}(1)$ lookup performance and minimal memory footprint:

\begin{equation}
    h(\boldsymbol{p}) = \text{hash}\left(\left\lfloor \frac{x}{\delta} \right\rfloor, \left\lfloor \frac{y}{\delta} \right\rfloor, \left\lfloor \frac{z}{\delta} \right\rfloor\right) \,,
\end{equation}
\noindent where {\small$\left\lfloor \cdot \right\rfloor$} denotes the element-wise floor operation applied to each coordinate of the position vector {\small$\boldsymbol{p} = [x, y, z]\in\mathbb{R}^3$}, and $\delta$ is the spatial resolution parameter. The spatial grid is used as a pure mathematical coordinate system where memory allocation occurs only upon data storage. This enables:
\begin{itemize}
    \item {Spatial coverage}: Any 3D position can be mapped to a cell without boundary constraints. The model retains historical data even when the state space expands without the risk of reinitialization of temporal flow dynamics.
    \item {Memory efficiency}: Storage requirements are proportional to the visited locations, rather than the map bounds.
    \item {Dynamic scalability}: The representation adapts seamlessly to environments of arbitrary size.
\end{itemize}

\subsection{Global Temporal Model Architecture}

\begin{figure}[]
    \centering
    \begin{subfigure}[b]{\linewidth}
        \centering
        \includegraphics[width=\linewidth]{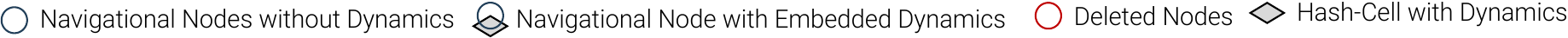}
        \label{fig:top_wide}
    \end{subfigure}
    \begin{subfigure}[b]{0.5\columnwidth}
        \centering
        \includegraphics[width=0.8\linewidth]{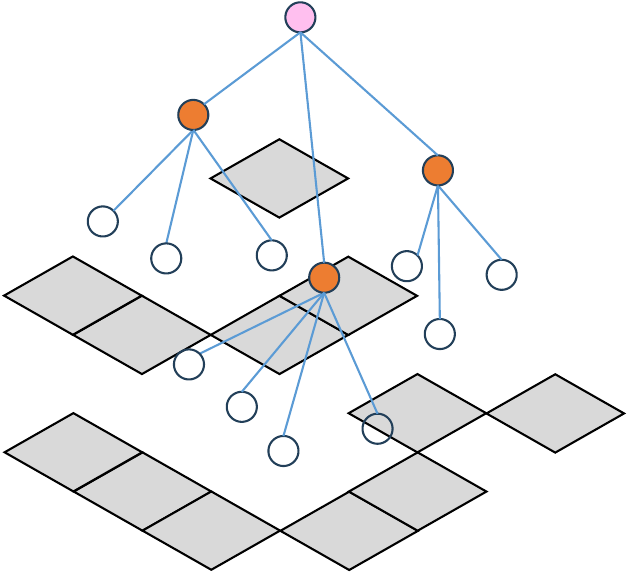}
        \caption{Before Binding (Hash Storage)}
        \label{fig:dynamics_in_cell}
    \end{subfigure}%
    \hfill
    \begin{subfigure}[b]{0.5\columnwidth}
        \centering
        \includegraphics[width=0.8\linewidth]{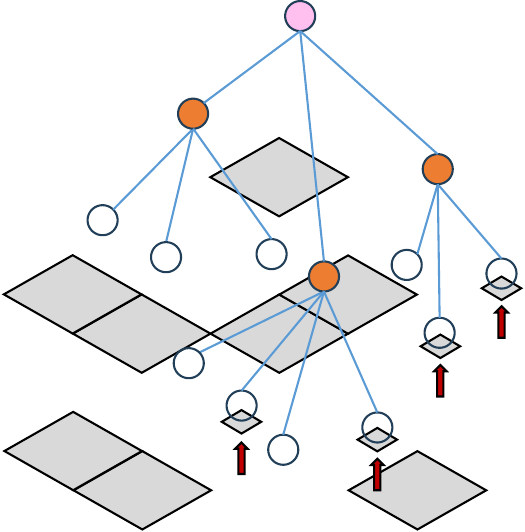}
        \caption{After Binding (Node Ownership)}
        \label{fig:dynamics_node}
    \end{subfigure}
    \begin{subfigure}[b]{0.5\columnwidth}
        \centering
        \includegraphics[width=0.8\linewidth]{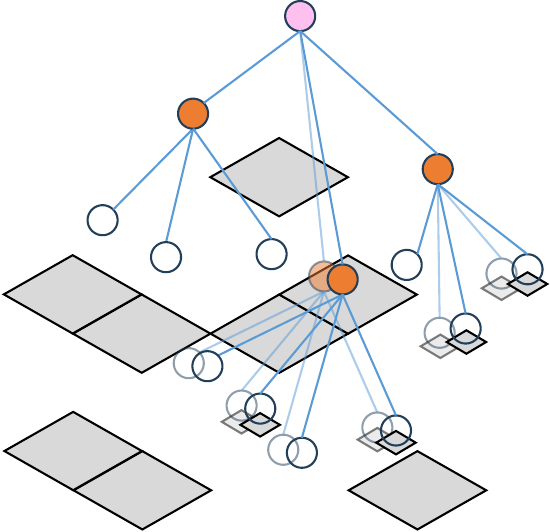}
        \caption{After Loop Closure}
        \label{fig:dynamics_move}
    \end{subfigure}%
    \hfill
    \begin{subfigure}[b]{0.5\columnwidth}
        \centering
        \includegraphics[width=0.8\linewidth]{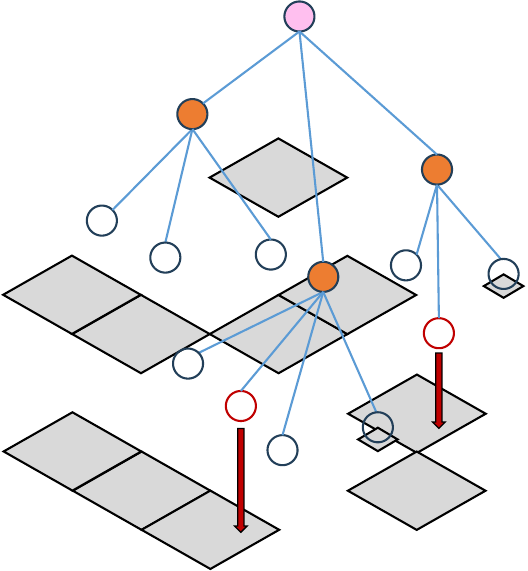}
        \caption{After Node Removal}
        \label{fig:dynamics_back}
    \end{subfigure}
    \caption{Illustration of temporal ownership transfer. (a) Temporal flow dynamics are first accumulated in spatial hash cells. (b) After a stability window, dynamics are bound to the nearest navigational node, which becomes the sole owner of the temporal history. (c) Following loop closure, nodes can be repositioned and with them the temporal flow dynamics. (d) If a node is removed during loop closure, the dynamics are restored into hash space and remain available for reassociation.}
    \label{fig:distribution_comparison}
\end{figure}

To support temporal reasoning over dynamic environments, Aion represents spatiotemporal flow dynamics directly on top of a hierarchical 3DSG. Each node in the graph corresponds to a navigational node and becomes a unit of temporal prediction. This allows the system to forecast human activity or motion trends in specific regions of the environment.

\subsubsection{Global Dynamics Model}

Rather than maintaining individual temporal models per node, Aion employs a single global temporal model that captures inter-node temporal dependencies while maintaining computational efficiency.

Given a dynamic entity located at position $\boldsymbol{p}_h^{}$ with orientation $\theta_h^{}$, the system associates it to the nearest node {\small$\boldsymbol{v}^*_{} \in \mathcal{V}^t_n$} in the navigational abstraction layer, \ie:
\begin{equation}
    \boldsymbol{v}^* = \argmin_{\boldsymbol{v}^t_{n,i} \in \mathcal{V}^t_n} \|\boldsymbol{x}^t_{n,i} - \boldsymbol{p}_h\|_2 \,, \quad \text{s.t.} \quad \|\boldsymbol{x}^t_{n,i} - \boldsymbol{p}_h\|_2 \leq d_{\text{max}} \,,
\end{equation}
\noindent where $\boldsymbol{x}^t_{n,i}$ is the node 3D position and $d_{\text{max}}^{}$ is a distance threshold. The orientation discretization process maps continuous orientation to discrete bins:
\begin{equation}
    \text{bin}(\theta_h^{}) = \left\lfloor \frac{\theta_h^{} + \pi}{2\pi/B} \right\rfloor \bmod B\,,
\end{equation}
\noindent with $B$ the number of orientation bins $b$ (typically 8, providing $45^{\circ}$ resolution).

In addition to per-node vectors, temporal activity across all spatial locations is tracked using a global vector defined over sparse hash keys:

\begin{equation}
\boldsymbol{s}(t) = [\boldsymbol{s}_{h_1}^{}(t), \boldsymbol{s}_{h_2}^{}(t), \ldots, \boldsymbol{s}_{h_N}^{}(t)] \,,
\end{equation}

\noindent where $h_1^{}, h_2^{}, \ldots, h_N^{}$ are spatial hash keys ordered deterministically.

This representation offers several advantages over grid-based MoDs \cite{kucner2023survey}. First, navigational nodes correspond to meaningful locations rather than arbitrary spatial cells and their density adapts to environmental structure rather than uniform discretization. Second, temporal patterns can be aggregated across the scene graph hierarchy with fewer spatial units compared to fine-grained grid representations.

\subsubsection{Long-term Consistency through Temporal Ownership Transfer}

Loop closure in graph-based mapping corrects accumulated drift by repositioning or removing navigational nodes. If temporal flow dynamics are stored only in spatial hash cells, such corrections can leave dynamics \textit{stranded} at outdated coordinates or duplicated across nodes, breaking temporal consistency.

To avoid this issue, Aion introduces a mechanism of temporal ownership transfer, in which temporal history is moved from spatial hash cells to navigational nodes once the pose graph has stabilized.
We define the binding function
\begin{equation}
    \phi: h_i^{} \mapsto \boldsymbol{v}_{n,i}^{} \,,
\end{equation}
\noindent which maps a spatial hash cell $c_{h_i^{}}^{}$ at key $h_i^{}$ to its owning navigational node $\boldsymbol{v}_{n,i}^{}$.

During exploration, temporal flow dynamics are first stored in hash space (\cref{fig:dynamics_in_cell}). After a stability window $\tau$ (a fixed time or an exploration horizon after which the graph structure is assumed to have converged, \eg no major pose graph updates or loop closures are expected), the temporal state vector $\boldsymbol{s}_{h_i^{}}^{}(t)$ associated with cell $c_{h_i^{}}^{}$ is transferred to the nearest node $v_{n,i}^{}$ (\cref{fig:dynamics_node}):
\begin{equation}
    \boldsymbol{s}_i^{}(t) \leftarrow \boldsymbol{s}_{h_i^{}}^{}(t), \quad \phi(h_i^{}) = \boldsymbol{v}_{n,i}^{} \,,
\end{equation}
\noindent and the data associated with hash cell $c_{h_i^{}}^{}$ is removed from the hash map. A lightweight redirect table maintains the mapping $\phi(c_{h_i^{}}^{})$, ensuring that any subsequent updates arriving at the coordinates of $h_i^{}$ are routed to node $v_{n,i}^{}$.
If the node is repositioned due to loop closure, its dynamics move with it, since $\boldsymbol{s}_i^{}(t)$ is stored directly in the node (\cref{fig:dynamics_move}). If the node is removed, ownership is released:
\begin{equation}
    \boldsymbol{s}_{h_i^{}}^{}(t) \leftarrow \boldsymbol{s}_i^{}(t), \quad \phi(h_i^{}) = \varnothing \,,
\end{equation}
\noindent so that the temporal history is rematerialized in hash space at the node’s final pose, ready to be reassociated with a new node when one appears nearby (Fig. \ref{fig:dynamics_back}).

This move-semantics approach prevents duplication of temporal histories, ensures memory efficiency, and preserves consistency under graph corrections. Hash cells act as provisional holders, while navigational nodes become the long-term owners of temporal flow dynamics once stable.

\begin{figure}[]
    \centering
    \includegraphics[width=\linewidth]{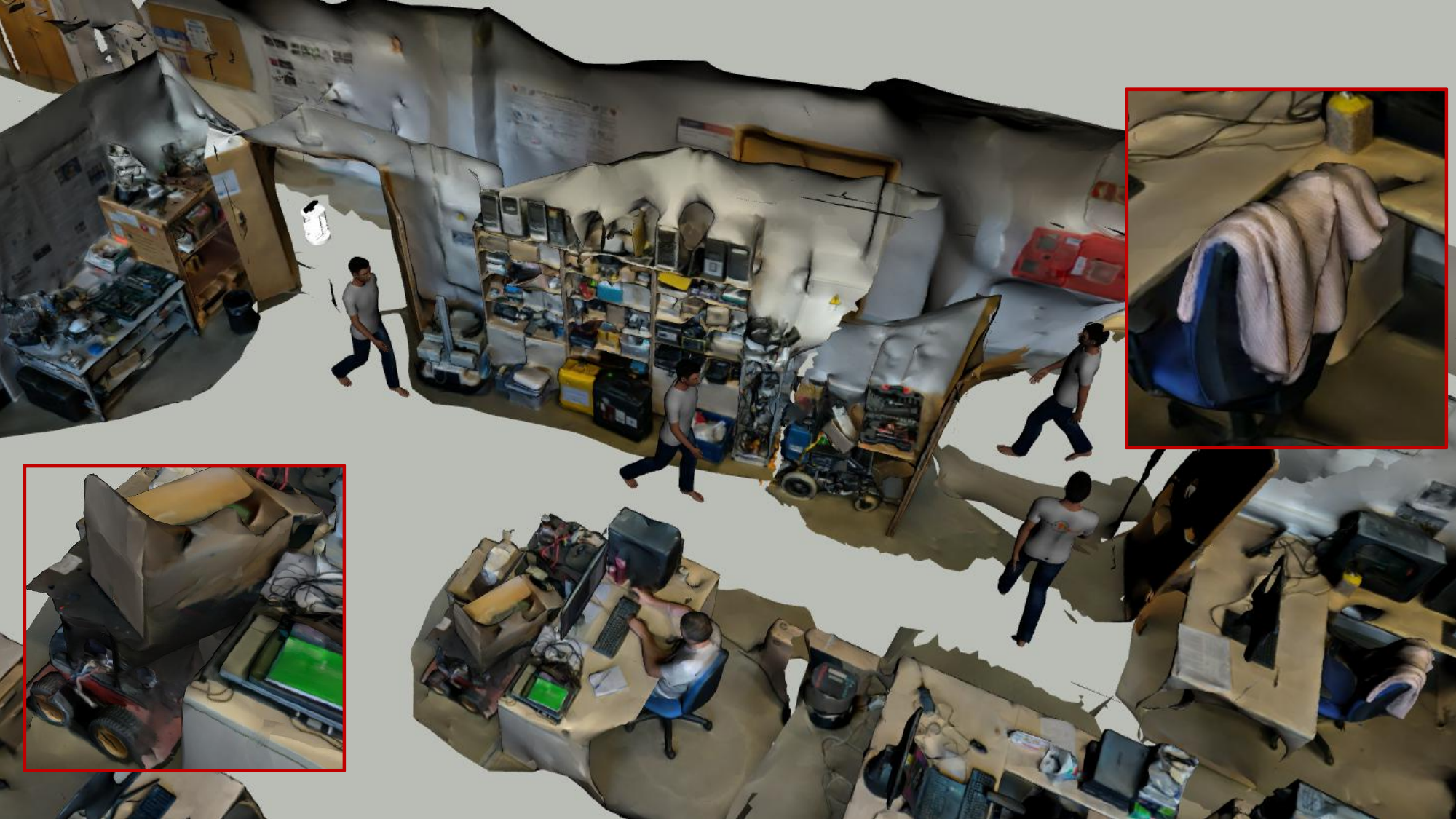}
    \caption{Virtual environment generated from real-world data and simulated agents.}
    \label{fig:simulation}
\end{figure}

\subsection{Real-time Integration Architecture}

Aion integrates seamlessly with existing 3DSG systems, specifically Hydra~\cite{hughes2022hydra}, through an architecture designed to preserve real-time performance without altering core scene graph processing. This is achieved through:
 \begin{itemize}
     \item \textbf{Asynchronous Processing}: Scene graph updates and agent detections are handled asynchronously, ensuring that temporal modeling does not block real-time 3DSG construction. 
     A dedicated temporal modeling thread accumulates observations and periodically updates both the global FreMEn and historical models.
    \item \textbf{Efficient Memory Management}: A sparse spatial hashing scheme bounds memory growth in proportion to environment coverage rather than overall map size, critical for long-term autonomous operation.
    \item \textbf{Service Interface}: Temporal predictions are exposed via ROS services, enabling integration with existing navigation and planning systems without requiring architectural changes to client systems. Aion extends Hydra’s capabilities by adding a parallel navigational layer that enriches the existing scene graph, providing additional temporal context.
 \end{itemize}
\section{Experimental Evaluation}

\subsection{Experimental Setup}

We evaluate the effectiveness of Aion in capturing and predicting motion dynamics in indoor environments.
Owing to the lack of directly comparable methods using 3DSGs for spatio-temporal dynamics modeling, we define a structured comparison against a grid-based approach (based on~\cite{molina2018modelling}), which serves as the reference baseline in our analysis.

\subsubsection{Dataset and Environment}

We developed a synthetic dataset derived from a virtual reconstruction of our university office environment, comprising a central workspace and a connected hallway (\cref{fig:simulation}).The dataset consists of 20 distinct scenes, each featuring 6 human agents navigating the environment along predefined routes. In each scenario, a mobile robot traverses the environment following specified paths. We use the simulator-provided detections to compute the spatio-temporal model discussed in~\cref{sec:methodology}.

\subsubsection{Comparison Framework}

Our method employs a sparse graph-based hybrid representation, where spatial locations are encoded as nodes in a 3DSG and hash cells. Given the mismatch in representation granularity and topology, a direct comparison is non-trivial. To enable a fair evaluation, we propose a two-stage comparison procedure. First, we construct a reference grid-based temporal model, which aggregates movement patterns into fixed-size spatial cells. Second, we discretize the output of our method into the same grid resolution, allowing for a direct comparison of the modeled dynamics.

We define two categories of evaluation, focusing on both observed and predicted spatio-temporal dynamics:
\begin{itemize}
    \item \textbf{Historical Dynamics}: This evaluation assesses how well the method captures aggregate movement patterns from past observations.
    \item \textbf{Temporal Predictions}: In this setting, we evaluate the ability of the method to capture cyclical dynamics pattern using a learned temporal model. Specifically, we use the Fremen temporal model~\cite{krajnik2017fremen} in both the grid-based and graph-based representations inspired by Molina~\etal~\cite{molina2018modelling}.
\end{itemize}

\begin{figure}[]
    \centering
        \begin{subfigure}[b]{\columnwidth}
        \centering
        \includegraphics[width=\textwidth]{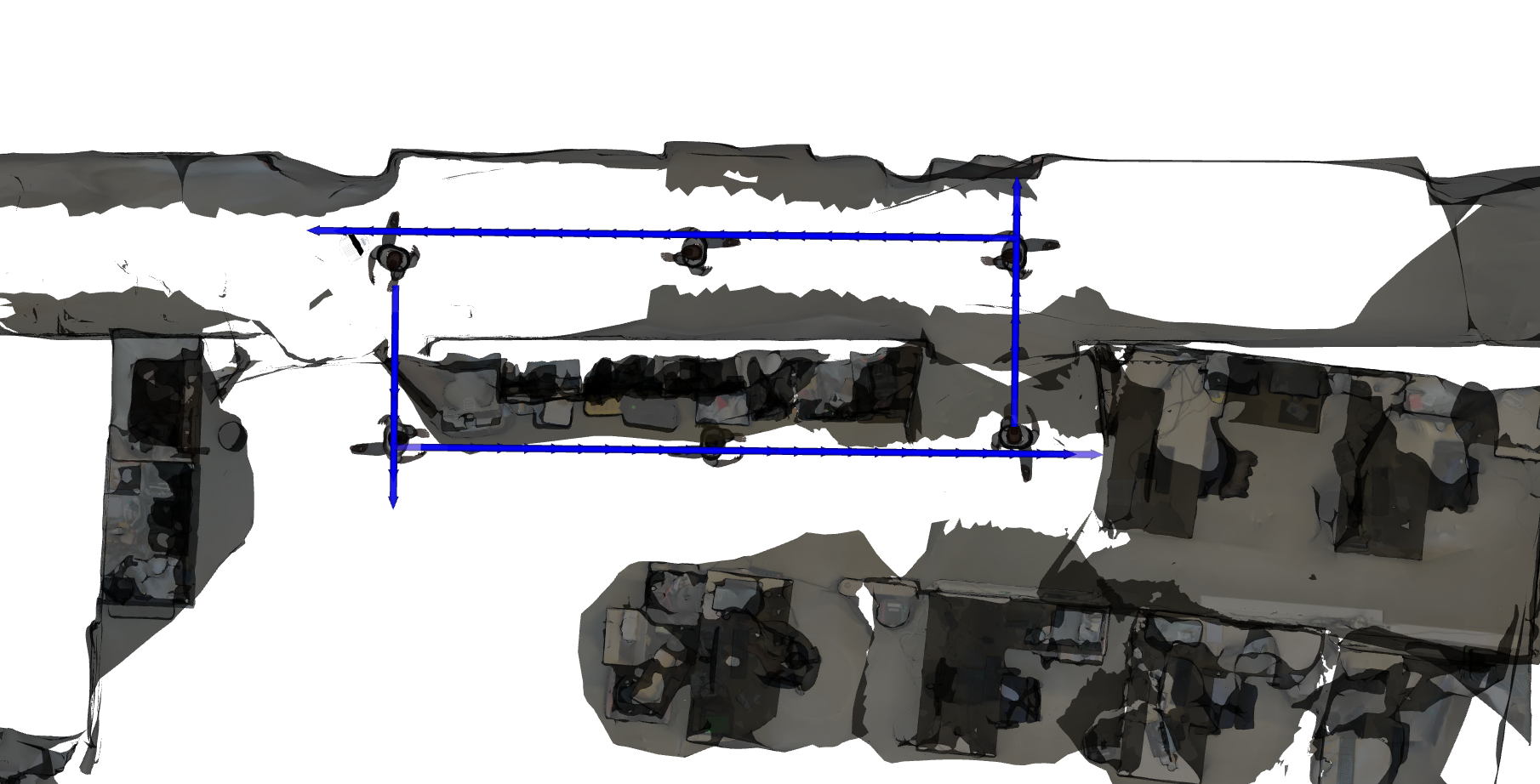}
        \label{fig:pattern_comparison_grid}
    \end{subfigure}%
    \hfill
    \begin{subfigure}[b]{\columnwidth}
        \centering
        \includegraphics[width=\textwidth]{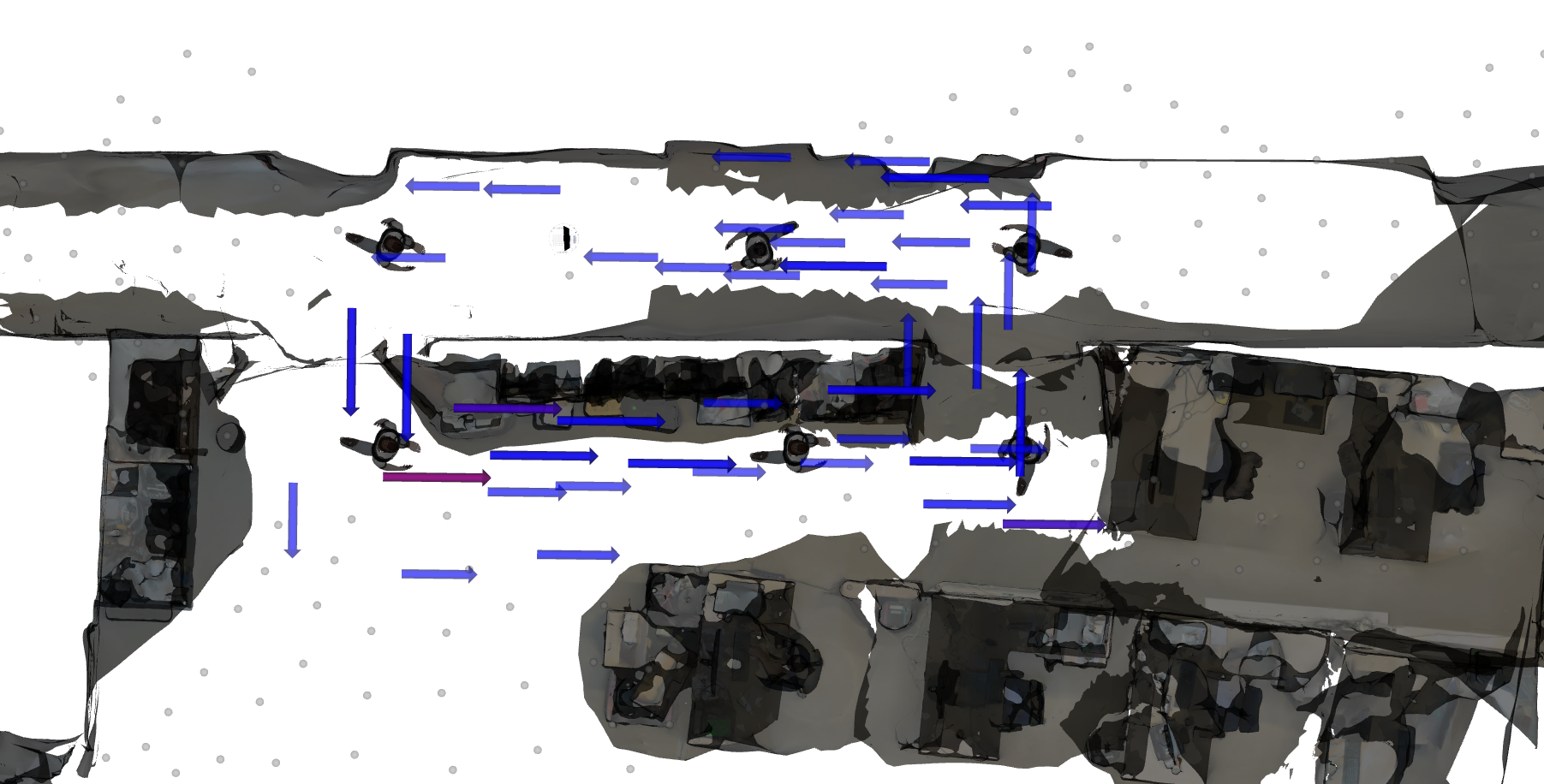}
        \label{fig:pattern_comparison_aion}
    \end{subfigure}

    \caption{Qualitative comparison of motion dynamics reconstructed for one scene with the grid-based system (top) and Aion (bottom). Arrow markers indicate the dominant direction of motion. Despite the different spatial discretization, Aion is capable of reproducing directional patterns.}
    \label{fig:distribution_comparison}
\end{figure}

\subsubsection{Evaluation Metrics}

Given the probabilistic and directional nature of the data, we employ the following metrics to compare the resulting distributions:

\begin{itemize} 
    \item \textbf{Jensen-Shannon (JS) Divergence} $(0-1)$: Measures the similarity between two probability distributions, bounded between 0 (identical) and 1 (maximally different). Suitable for general-purpose distribution comparison.
    \item \textbf{Bhattacharyya Distance} $(0-\infty)$: Quantifies probability mass overlap between distributions. Higher values indicate less overlap in the probability densities.
    \item \textbf{Wasserstein Distance}: Also known as Earth Mover's Distance, this metric reflects the cost of transforming one distribution into another. For directional data, it corresponds to the average angular displacement required.
    \item \textbf{Circular Correlation}: Directional correlation between two vector fields.
\end{itemize}

\subsection{Results and Analysis}

To assess similarity between the graph-based and grid-based models, we compute metrics per scene and report the mean and standard deviation across all datasets (see~\cref{tab:metrics}). This allows us to capture both the central tendency and variability of each method's performance across different environments. 

The results indicate that both methods capture related aspects of the underlying dynamics, though the degree of alignment varies across measures. For entropy, the models show consistent agreement in the historical case, with low Wasserstein distances (W = $0.12$) suggesting that both representations identify comparable levels of variability in human activity. Predictions also show overlap (JS = $0.35$), but with higher variance, suggesting that cyclical temporal models are more sensitive to how the spatial representation aggregates data. Flow magnitude distributions are less closely aligned: while historical activity levels show moderate similarity (JS = $0.53$), predicted flows diverge more strongly (JS = $0.68$), pointing to differences in how each representation scales activity over time.

Directional flow analysis produces the largest quantitative discrepancies, with low circular correlations (r = $0.12$) and high Wasserstein distances. However, qualitative comparisons (\cref{fig:distribution_comparison}) demonstrate that directional patterns are in fact well reproduced. The gap arises primarily from differences in spatial discretization: the grid-based system evaluates orientation over fixed cells, while Aion employs an adaptive structure of nodes and hash-cells whose placement varies across runs and datasets. This flexibility allows Aion to capture meaningful motion at navigationally relevant locations, but it also introduces misalignments when distributions are forced into a grid-based evaluation framework.

\begin{table*}
\centering
\caption{Comparison between Aion and grid-based structure for different data types. The results in the table show that comparable accuracies are obtainable with Aion, with the additional benefits of scale through hierarchy as well as data structure optimizations.}
\label{tab:metrics}
\begin{tabular}{llccccc}
\toprule
Map Type & Data Type & JS Divergence & Bhattacharyya Distance & Wasserstein Distance & Circular Correlation\\
\midrule
Entropy & Historical & 0.49 $\pm$ 0.13 & 1.3 $\pm$ 0.68 & 0.12 $\pm$ 0.05 & --\\
Entropy & Predicted & 0.35 $\pm$ 0.32 & 3.83 $\pm$ 4.69 & 0.18 $\pm$ 0.14 & --\\
Flow & Historical & 0.53 $\pm$ 0.11 & 2.02 $\pm$ 2.06 & 40.17 $\pm$ 21.82 & --\\
Flow & Predicted & 0.68 $\pm$ 0.02 & 6.42 $\pm$ 3.76 & 76.16 $\pm$ 16.93 & --\\
Direction & Historical & -- & -- & 15 $\pm$ 11.23 & 0.12 $\pm$ 0.07 \\
Direction & Predicted & -- & -- & 103.03 $\pm$ 14.15 & 0.1 $\pm$ 0.05 \\
\bottomrule
\end{tabular}
\end{table*}

\subsection{Planning}

\begin{figure}
    \centering
    \includegraphics[width=\linewidth]{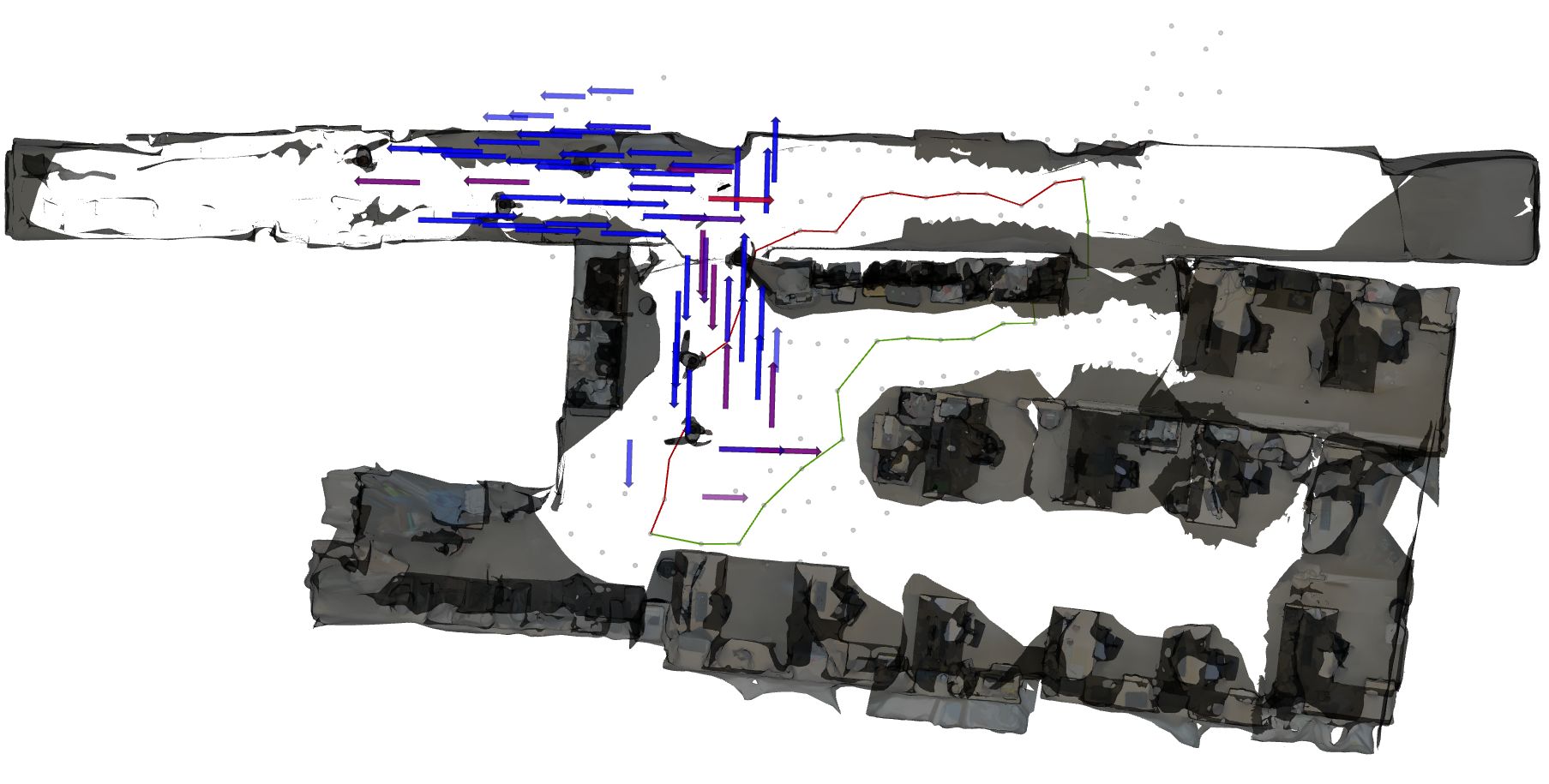}
    \caption{Qualitative comparison of $A^*_{}$ planning results on the navigational layer of the 3DSG. In red: the baseline planning without temporal dynamics. In green: the computed path obtained with our proposed method, which integrates temporal motion patterns as edge costs. The resulting path demonstrates improved efficiency avoiding crowded areas.
    }
    \label{fig:planning}
\end{figure}

This section illustrates how the temporal flow dynamics encoded in Aion can be used to inform navigation decisions. The goal is not to propose a new planning algorithm, but to demonstrate that the additional information provided by Aion can be integrated into standard methods such as $A^*_{}$.

Each node in the navigational graph is annotated with dynamic attributes (\textit{entropy}, \textit{flow magnitude}, and \textit{flow direction}) combined into a temporal cost reflecting both the variability of activity in a region and whether a planned movement aligns with the dominant flow. Edges that pass through uncertain or opposing-flow regions therefore incur a higher cost than those through stable or aligned areas. The resulting traversal cost between nodes $i$ and $j$ is defined as:
\begin{equation}
    \text{cost}(i, j) = d(i, j) + \left( \bar{c}^{}_t + c^{}_d(i, j) \right) \cdot d(i, j)\,,
\end{equation}
where $d(i, j)$ is the Euclidean distance between nodes $i$ and $j$, $\bar{c}_t$ represents the average temporal cost of the two nodes, and $c^{}_d(i, j)$ is a directional penalty. This formulation extends conventional distance-based planning with dynamic penalties that encourage safer and more context-aware paths.

\textit{Results.} As shown in~\cref{fig:planning}, paths planned with Aion’s temporal dynamics tend to avoid regions of high entropy and flow dynamics.
In contrast, a purely distance-based $A^*_{}$ planner often selects shorter but less robust routes that cut directly through uncertain areas. This qualitative comparison highlights that Aion’s representation can provide meaningful guidance for navigation in dynamic environments, even when applied within standard planning frameworks.
\section{Conclusion}\label{sec:conclusion}

We introduce Aion, a system that augments hierarchical 3D scene graphs with temporal flow dynamics, providing a step toward semantically informed 4D scene representations.  Our approach embeds dynamics into 3DSGs, representing the first unification of Maps of Dynamics and Scene Graphs. Aion enables temporal reasoning at navigationally meaningful locations, and maintains consistency under evolving graph topologies through a mechanism of temporal ownership transfer, ensuring that motion histories remain valid under structural updates of the graph (e.g., during optimization steps such as loop closure).

Our experimental results demonstrate that Aion captures and reproduces spatio-temporal motion patterns with good agreement to grid-based baselines, while offering a more interpretable and semantically structured representation. With this initial version, we provide a complete implementation and integration into Hydra's 3DSG framework, laying the groundwork for future research on applying 4D scene graphs in human-aware navigation, advancing long-term autonomy in highly dynamic settings.

Overall, this work lays the foundation for future research on applying 4D scene graphs in real-world environments. Future work will be directed towards larger-scale deployments, and human-aware navigation, to demonstrate the potential of Aion for long-term autonomy in highly dynamic settings.


\bibliographystyle{unsrt}
\bibliography{bibliography}

\end{document}